\documentclass{article}
\usepackage{ijcai19}

\usepackage{times}
\usepackage{xcolor}
\usepackage{soul}
\usepackage{url}
\usepackage[hidelinks]{hyperref}
\usepackage[utf8]{inputenc}
\usepackage[small]{caption}
\usepackage[utf8]{inputenc}
\usepackage{graphicx}
\usepackage{amsmath}
\usepackage{amssymb} 
\usepackage{booktabs}
\usepackage{amsfonts}
\usepackage{bm}
\usepackage{multirow}

\title{3D Graph Convolutional Networks with Temporal Graphs: A Spatial Information Free Framework For Traffic Forecasting}

 \author{
 Bing Yu \thanks{Equal contributions.}$^1$, 
 Mengzhang Li$^*$$^2$$^,$$^3$, 
 Jiyong Zhang$^5$, 
 Zhanxing Zhu \thanks{Corresponding author.}$^3$$^,$$^4$
 \\
 \affiliations
 $^1$School of Mathematical Sciences, Peking University, Beijing, China\\
 $^2$Academy for Advanced Interdisciplinary Studies, Peking University, Beijing, China\\
 $^3$Center for Data Science, Peking University, Beijing, China\\
 $^4$Beijing Institute of Big Data Research (BIBDR), Beijing, China\\
 $^5$School of Automation, Hangzhou Dianzi University, Hangzhou, China\\
 \emails
 \{byu,mcmong,zhanxing.zhu\}@pku.edu.cn 
 }

\begin{document}
\maketitle
\begin{abstract}
Spatio-temporal prediction plays an important role in many application areas especially in traffic domain. However, due to complicated spatio-temporal dependency and high non-linear dynamics in road networks, traffic prediction task is still  challenging. Existing works either exhibit heavy training cost or fail to accurately capture the spatio-temporal patterns, also ignore the correlation between distant roads that share the similar patterns. In this paper, we propose a novel deep learning framework to overcome these issues: 3D Temporal Graph Convolutional Networks (3D-TGCN). Two novel components of our model are introduced. (1) Instead of constructing the road graph based on spatial information, we learn it by comparing the similarity between time series for each road, thus providing a spatial information free framework. (2) We propose an original 3D graph convolution model to model the spatio-temporal data more accurately. Empirical results show that 3D-TGCN could outperform state-of-the-art baselines. 
\end{abstract}

\section{Introduction}
Traffic speed prediction is a crucial task for many key purposes in intelligent traffic systems and urban planning. For example, it is useful for not only explicit tasks such as calculating how many lanes a road should have, monitoring whether some places have a traffic jam, but it can also reflect road conditions for downstream traffic problems, e.g., employing it as an important feature for estimating time of arrival, route planning and traffic light control.

In traffic forecasting problems, we typically choose density \cite{kriegel2008statistical}, speed \cite{ma2015long} and volume \cite{okutani1984dynamic} as indicators to characterize current traffic conditions. The traffic forecasting problem can be categorized into three types, namely, based on the length of prediction, i.e., short-term (less than 30 min) \cite{vlahogianni2005optimized} and long term(30 $\sim$ 60 min) \cite{ostring2001influence}, based on the data source, i.e., fixed sensors on several roads \cite{li2018diffusion} and moving GPS trajectories treated with map-matching algorithm \cite{castro2012urban}, and based on the road type, i.e., urban road \cite{stathopoulos2003multivariate} and highway \cite{fitzpatrick2000speed}. These prediction types are  challenging due to the complexity of spatio-temporal dependencies and particularly the uncertainty of long-term forecasting. 

Before data-driven approaches spring up, researchers usually apply mathematical tools such as differential equations and traditional traffic knowledge to simulate traffic behaviour by numerical simulation~\cite{vlahogianni2015computtionl}. This makes strong assumptions, such as drivers' identical behaviour and no sudden accidents. In the past several decades, many statistical and machine learning methods such as Auto-Regressive Integrated Moving Average (ARIMA) models \cite{yu2004switching,williams2003modeling}, support vector regression (SVR) \cite{hong2011traffic}  
were proposed. However, these methods rely on the stationary assumption of time series that are hard to model highly non-linear traffic flow and they ignore the correlation between different roads. Meanwhile, some works consider spatial structure of input data, namely, applying convolutional neural network (CNN) to capture the adjacent correlation and recurrent neural network (RNN) or long short-term memory (LSTM) network on time axis  \cite{ma2017learning,wu2016short,zhao2017lstm}. However, normal convolutional operation applies on grid structures such as images and videos, not suitable for traffic networks; and training of RNN, LSTM networks is time consuming and difficult.

To model temporal pattern and spatial dependencies effectively, recent works introduce graph convolutional network (GCN) to learn the traffic networks~\cite{li2018diffusion,defferrard2016convolutional}. DCRNN~\cite{li2018diffusion} utilizes the bi-directional random walks on the traffic graph to model spatial information; and captures temporal dynamics by gated recurrent units (GRU). This sequence-to-sequence model performs well at the cost of very expensive computation during training. STGCN \cite{yuspatio} relies on graph convolution on spatial domain and 1-D convolution along time axis. Though STGCN could significantly save training time due to its pure convolution operations, it processes graph information and time series separately, unfortunately, which might ignore accurately modeling the interaction between spatial and temporal dynamics. 
 
 On the other hand, existing graph-based prediction approaches consider the relationship between roads by  relying on   the graph constructed based on the spatial distance (e.g. GPS distance), or road connectivity. However, in some practical scenarios, the spatial adjacency matrix is difficult to generate, since for some free editable maps such as OpenStreetMap \cite{haklay2008openstreetmap}, acquiring up-to-date and  accurate spatial topology information is hard. Meanwhile, the service of commercial map is expensive and its API will constrain query times for distance calculation\footnote{For example, Baidu Map, one of the biggest commercial map app around the world, provides individual developers with at most 30, 000 query times per day and its full basic service costs 10 thousand dollars per month. See \url{http://lbsyun.baidu.com/apiconsole/auth/privilege}.}.
  More importantly, we argue that this way of graph construction unfortunately ignores the \emph{correlation between distant roads that share the similar temporal pattern}.  For instance, at rush hours, most roads near office buildings that have similar traffic patterns will encounter traffic jams in the same period. Both of these influence could be extracted from the time series themselves.

\begin{figure}
	\begin{minipage}[t]{0.5\linewidth}
		\centering
		\includegraphics[width=1.7in]{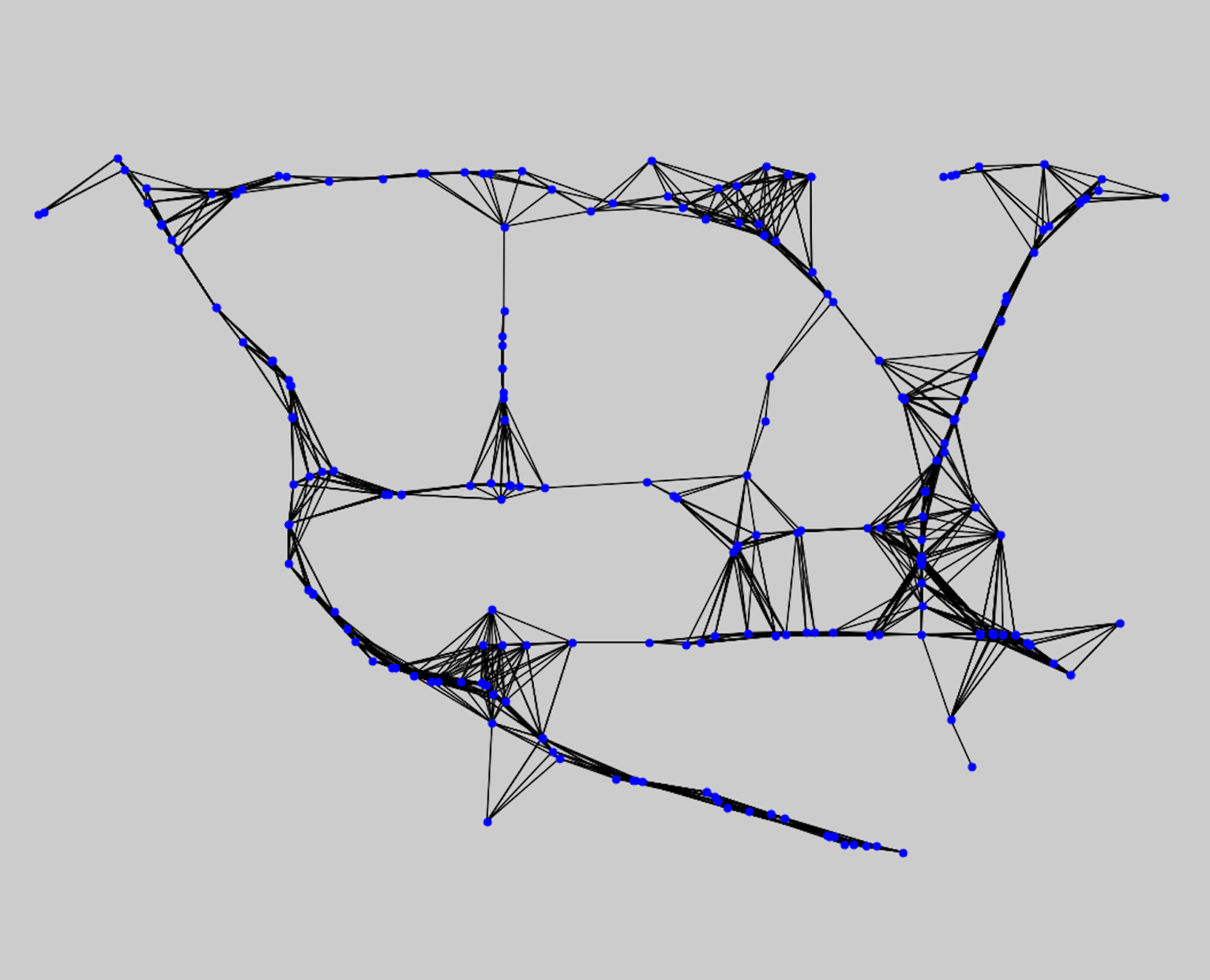}
	\end{minipage}%
	\begin{minipage}[t]{0.5\linewidth}
		\centering
		\includegraphics[width=1.7in]{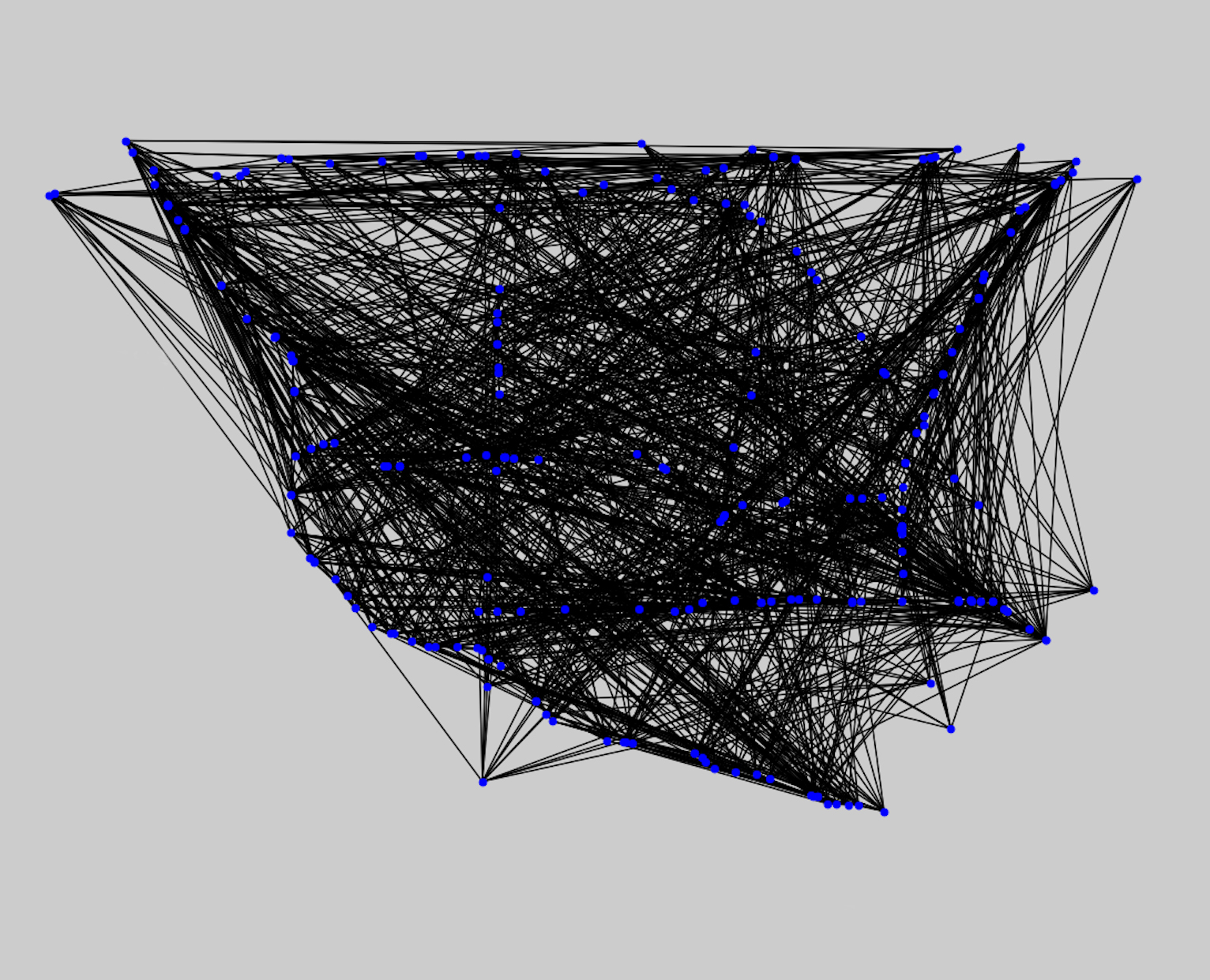}
	\end{minipage}
	\caption{\label{fig:stg} Comparison between the constructed graphs based on spatial distance and similarity of time series, respectively.}
\end{figure}

To overcome the drawbacks above, we propose a novel methodology for improving traffic prediction from aspects of both model design and graph construction. To  extract  better spatio-temporal dependencies, we propose a 3D graph convolution network  where 3D convolution is applied to simultaneously learn the spatial and temporal patterns together. Furthermore, we offer a spatial information free approach for constructing the graph for traffic network, purely relying on the similarity of time series for each road. This new proposal could capture more effective patterns between different roads than the spatial graph, facilitating superior prediction performance.   
The contributions of this work can be summarized as follows.  
\begin{enumerate}
\item We create a 3D GCN model to jointly learn the static road graph and temporal dynamics together. This new network structure strikes a better balance between training efficiency and effectiveness of feature learning.
\item Instead of using spatial information, we construct the adjacency matrix between nodes only according to the time series similarity  by dynamic time warping (DTW) algorithm. The difference between the two types of graph construction is presented in Figure~\ref{fig:stg}. It solves the difficulty of acquirement of geographic information. We empirically show that the performance of this temporal graph performs much better than spatial graph. 
 To the best of our knowledge, it is the first time to put aside spatial adjacency matrix and construct spatio-temporal graph by a data-driven method which extracts effective features from road networks' time series themselves.
\item We conduct extensive experiments on two open large-scale real-world datasets. Results show both of 3D GCN model and our spatial information free graph obtains significant improvement over state-of-the-art baseline methods.
\end{enumerate}

\section{Preliminary}
\subsection{Traffic Forecasting Problem}
We can represent the road network as a graph $\mathcal{G} = (\mathcal{V}, \mathcal{E}, W)$, where $\mathcal{V}$ is a finite set of nodes $\vert \mathcal{V}\vert = n$, corresponding to observation of $n$ sensors or roads; $\mathcal{E}$ is a set of edges and $W \in \mathbb{R}^{n \times n}$ is a weighted adjacency matrix representing the nodes proximity (e.g. spatial distance or temporal similarity). Denote the observed graph signal $X \in \mathbb{R}^{n\times d}$, the element of which means observed traffic flow of each sensor. Let $X^{(t)}$ represents the graph signal on time step $t$. The aim of traffic forecasting is learning a function $h(\cdot)$ from previous $M$ speed observations to predict next $H$-th traffic speed from $N$ correlated sensors on the road network. 
\begin{equation}
[X^{(t-M+1)}, \cdots, X^{t}] \xrightarrow[\mathcal{G}]{h(\cdot)} [ X^{t+H}]
\end{equation}
\subsection{Convolution on graphs}
Different from normal convolutional operation which processes regular grids on images or videos, graph convolution operation mainly has two types. One is based on the spectrum of the graph Laplacian, namely, extending convolutions to graphs in spectral domain by finding the corresponding Fourier basis~\cite{bruna2013spectral}. The other is generalizing spatial neighbours by rearranging the neighbours of vertices in a graph to apply regular convolutional operation~\cite{niepert2016learning}. 

Graph convolutional operation based on the spectrum is able to extract local features with different reception fields from non-Euclidean structures\cite{hammond2011wavelets}. It is defined over a graph $\mathcal{G} = (\mathcal{V}, W)$, where $\mathcal{V}(\vert \mathcal{V} \vert = n)$ is the set of all vertices in this graph and $W \in \mathbb{R}^{n \times n }$ is the adjacency matrix whose entries represent certain distance between vertices. Let its normalized graph Laplacian matrix be $L = I_n - D^{-\frac{1}{2}} W D^{-\frac{1}{2}} = U \Lambda U^T$, where $I_n$ is an identity matrix, $D \in \mathbb{R}^{n\times n}$ is the degree matrix with $D_{ii} = \sum_{j}W_{ij}$. $U \in \mathbb{R}^{n\times n}$ is the Fourier basis which is composed of eigenvectors of Laplacian matrix $L$. The graph signal $X \in \mathbb{R}^n$ is filtered by a diagonal matrix kernel $\Lambda_{\theta}$ with multiplication between $U$ and $U^T X$:
\begin{equation}
{\hat{X}} =   U   \Lambda_{\theta} U^T {X}
\label{gcn_1st}
\end{equation}
where the kernel $\Lambda_{\theta}$ is a group of parameters to be trained,  and ${\hat{X}}$ denotes the output of this GCN layer.

To reduce the number of parameters and generate a kernel which has better spatial localization, the kernel $\Lambda_{\theta}$ can be redesigned as the Chebshev polynomial $\Lambda_{\theta} \approx \sum_{k=0}^{K-1} \theta_k P_k(\tilde{\Lambda})$, It has a truncated order $K-1$ and utilizes the largest eigenvalue of $L$ to rescale $\Lambda$:  $\tilde{\Lambda} = 2 \Lambda / \lambda_{max} - I_{n}$\cite{defferrard2016convolutional}. 

Then we could reformulate Equation \ref{gcn_1st} into:

\begin{equation}
{\hat{X}}  \approx \sum_{k=0}^{K-1} U \theta_k P_k(\tilde{\Lambda})U^T{X}=\theta_k P_k(\tilde{L}) {X}
\end{equation}
where $\tilde{L} = 2L/ \lambda_{max} - I_{n}$ is the scaled Laplacian and $\theta_1, \theta_2, \cdots, \theta_k$ are parameters which could be trained by Back Propagation.

\subsection{Similarity of Temporal Sequences}
 Generally speaking, the methods for measuring the similarity between time series can be divided into three categories: (1) timestep-based, such as Euclidean distance reflecting point-wise temporal similarity; (2) shape-based, such as Dynamic Time Warping~\cite{berndt1994using} according to the trend appearance; (3) change-based, such as Gaussian Mixture Model(GMM)\cite{povinelli2004time} which reflects similarity of data generation process. 

In this work, we utilize Dynamic Time Warping to measure similarity  i.e., the spatial shape of time series, between different roads to predict future time series. Given two time series $X = (x_1, x_2, \cdots, x_n)$ and $Y = (y_1, y_2, \cdots, y_m)$ whose length are $n$ and $m$. We first introduce a series distance matrix $M_{n\times m}$ whose entry is Euclidean distance of two series points $M_{i, j} = \vert x_i - y_j \vert$. Then we can define the cost matrix (accumulated distance matrix) {$M_c$}:
\begin{small}
\begin{equation}
M_c(i,j) = M_{i,j} + \min(M_c(i, j-1), M_c(i-1, j), M_c(i,j))\label{mc}
\end{equation}
\end{small}

After several iterations of $i$ and $j$ (i.e., each of them increases from 1 to $n$ and $m$), $M_c(n, m)$ is the final distance between $X$ and $Y$ with the best alignment which can represent the similarity between two time series.

From Equation \ref{mc} we can tell that Dynamic Time Warping is an algorithm based on dynamic programming and its core is solving the warping curve, i.e., matchup of series points $x_i$ and $y_j$. In other words the "warping path"
\begin{align}
\Omega = (\omega_1, \omega_2, \cdots, \omega_K), \quad \max(n, m) \leq K \leq n+m\notag
\end{align}
is generated through iterations of Equation \ref{mc}. Its element $\omega_k = (i, j)$ means matchup of $x_i$ and $y_j$. The warping path $\Omega$ starts from $\omega_1 = (1, 1)$ and ends with $\omega_K = (n, m)$ thus every series points of $X$ and $Y$ must appear in $W$. Moreover, $i$ and $j$ in $\omega(i ,j)$ must increase monotonically to avoid crossover of each matchup. For instance, given $\omega_k = (i ,j)$ and $\omega_{k+1} = (i^\prime, j^\prime)$ then $ i\leq i^\prime \leq i+1$ and $ j\leq j^\prime \leq j+1$.

\section{Proposed Model: 3D-TGCN}
In this section, we explicitly formalize the spatio-temporal traffic prediction problem and describe our \textit{3D Temporal Graph Convolutional Networks}.

\subsection{Graph Generation}
Different from those proposed models that requires spatial adjacency matrix, 3D-TGCN could learn those roads' interior temporal pattern by calculating their corresponding time series' distance. This way of graph construction is completely data-driven, helping to capture more effective information than the priori given spatial information. For instance, if traffic data are aggregated every 5 minutes then each road has 288 time steps in one day. Given time series $A \in \mathbb{R}^{288\times 1}$ for one road and time series $B \in \mathbb{R}^{288\times 1}$ of another, then we could utilize Dynamic Time Warping algorithm to find optimal match and calculate distance of their time series.

\begin{figure}
\centering
\includegraphics[height=2in]{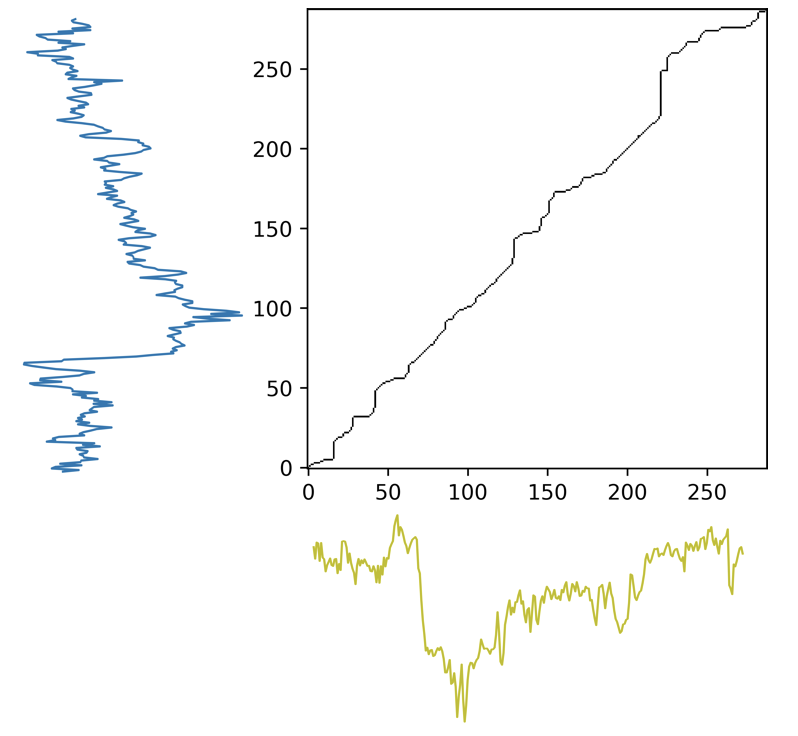}
 \caption{\label{fig:dtw}Two time series of different roads in one day and their warping path calculated by Dynamic Time Warping algorithm.}
 \end{figure}

As shown in Figure \ref{fig:dtw}, given two roads' time series whose length is 288 then we could achieve their warping path. The distance of those two time series could be calculated by Equation \ref{mc} (i.e., $M_c(288, 288)$ in this case). From the figure we could tell the warping path elongates along the diagonal since the trend of two time series are similar, consequently the difference between match $i$ and $j$ of the element $\omega_k = (i, j)$ of warping path $\Omega$ are close.

Then we generate topology network $W$. For each road $i$, we pick up its top $5\%$ most similar roads $\mathbb{S}=\{j, k, \cdots \}$ and let $W_{ij} = W_{ik} = \cdots = 1 $ while others $W_{is} = 0, s \notin \mathbb{S}$. Moreover, it is possible that $W_{ij} = 1$ while $W_{ji} = 0$, then we reassign $W_{ji} = 1$ if $W_{ij} = 1$. After this treatment, the constructed $W$ could be applied in our 3D-TGCN model, described in the following.

\subsection{3D Graph Convolution Networks}

\begin{figure*}[ht!]
\centering
\includegraphics[width=7in]{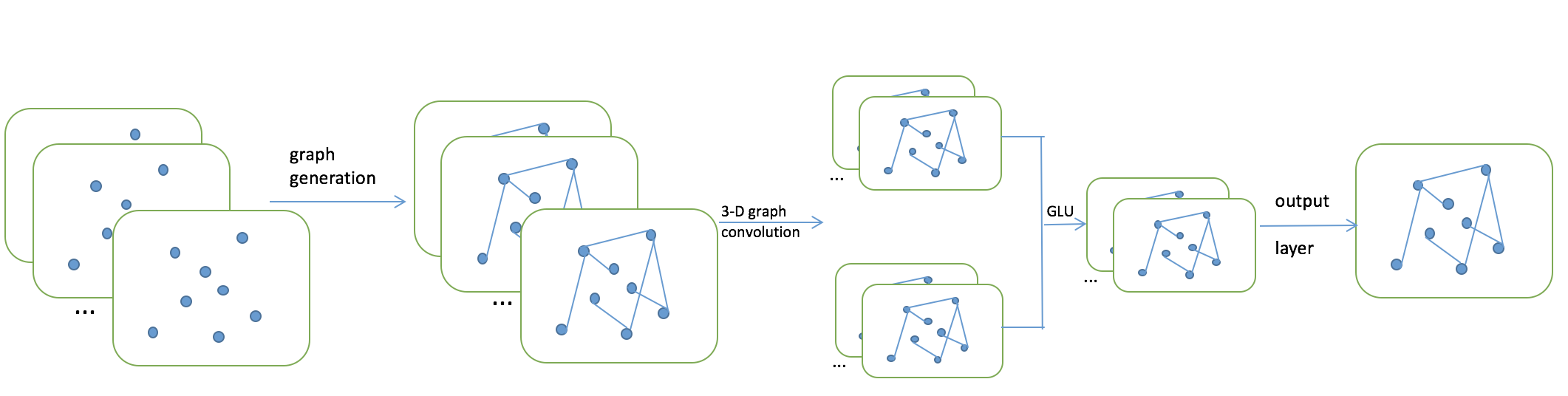}
\caption{\label{fig:tgcn}  Network architecture of 3D-TGCN}
\end{figure*}

\textbf{3D Graph Convolutional Layer}

Many existing approaches deal with  spatial and temporal dependencies separately since they utilize graph convolution on spatial dependencies and leverage 1-D CNN~\cite{yuspatio} or RNN-based models~\cite{li2018diffusion} to extract temporal dependency along time axis. For instance, if 1-D CNN was deployed in the temporal direction, the output of each 1-D convolution could be rewritten as,

\begin{equation}
\hat{X}_t =\sum_{t^{\prime} = 0}^{K_t -1} \theta_{t^{\prime}}^{(t)} \sum_{k=0}^{K-1} \theta_k  \tilde{L}^k X_{t - t^{\prime}}\label{g1}
\end{equation}
where $K_t$ is the size of convolutional kernel on time-axis at time step $t$.

We now propose a 3D graph convolutional operation on all dimensions, including graph topology and temporal direction.

For the input ${X}_t$ ($1\leq t\leq M$) with $C_i$ channels, it can be extended to multi-dimensional arrays $\mathcal{X}_t \in \mathbb{R}^{n \times C_i}$. The 3D graph convolutional layer integrates all dimensions together: 

\begin{equation}
\begin{aligned}
{\hat{X}}_{t, C_o}  = \sum_{i=1}^{C_i} \sum_{t^{\prime} = 0}^{K_t -1} \sum_{k=0}^{K-1}  \theta_{i, C_o, k, t^{\prime}} \tilde{L}^k X_{(t - t^{\prime}), i}\label{g2},\\
\quad t=K_t,K_t+1,...,M
\end{aligned}
\end{equation}
where $C_i$ and $C_o$ are the size of input and output of this 3D  graph convolutional layer,  respectively and $\theta_{i, C_o, k, t^{\prime}}$ is the parameter to be trained in each output channel of this layer. From Equation \ref{g2}, the graph convolution operator of each layer could be denoted as "$\Theta_{*\mathcal{G}} {X}$" with $\Theta \in \mathbb{R}^{C_i \times C_o \times K_t \times K}$. 

The 3D graph convolutional layer scans $K_t$ neighbours on time-axis without padding and $(K - 1)$-order neighbourhood of temporal graph $\mathcal{G}$ at the same time. This method shortens the length of sequences by $K_t -1$ each time. It follows by a gated linear units (GLU) whose input is: $[G, H] \in \mathbb{R}^{(M - K_t +1) \times (2\times C_o)}$ where $G, H$ is split in half with the size of $C_o$ channels. As a result, the final output of 3D graph convolutional layer is $ \hat{{X}} =  G \odot \sigma(H) \in \mathbb{R}^{(M-K_t+1) \times C_o}$ where $\odot$ denotes the Hadamard product and $\sigma(\cdot)$ denotes the sigmoid function.

This integrated design of 3D graph convolution allows us 
to jointly learn graph structure and temporal dynamics as a whole. It is also easy for building such multi-layer 3D graph convolutional structures.

\subsection{The Entire Architecture of 3D-TGCN Network}
Figure~\ref{fig:tgcn} sketches the overall architecture of our proposed  3D-TGCN model. It consists of four 3D graph convolutional blocks (3D-Conv blocks), one output block. Each 3D-Conv block contains two 3D graph convolutional layers and a layer normalization layer to prevent overfitting. The output block consists of several 3D graph convolutional layers or 1D temporal convolutional layers and a weight sharing fully-connected output layer to obtain the prediction $\hat{X}_{T+H} \in \mathbb{R}^n$. 

The $L_2$ loss and $L_1$ loss will be used together to train our model and the loss function of 3D-TGCN model could be formulated as below:
\begin{align}
L(\hat{X}; \Delta_{\theta}) =& \sum_{t}  (\frac{1}{2} \Vert \hat{X}_{t+H}-X_{t+H}) \Vert_2^2\notag\\
& + \Vert \hat{X}_{T+H}-X_{T+H} \Vert_1)
\end{align}

In summary, our 3D-TGCN model has several advantages:
\begin{itemize}
\item 3D-TGCN does not require  spatial adjacency matrix, instead, it constructs temporal adjacency matrix to learn temporal patterns of different roads in a pure data-driven way. 
\item The 3D graph convolution integrates all dimensions (i.e., time-axis on each road and correlation between different roads) into one graph convolutional networks. This design presents a better balance between training efficiency and effectiveness of feature learning on complex spatio-temporal graph, compared with STGCN and DCRNN. 
\item 3D-TGCN could be applied into many other tasks that have spatio-temporal features. Its universal framework can learn spatio-temporal dependencies between each participant. By calculating similarity between time series, 3D-TGCN could extract  important temporal pattern of different participants which might appear uncorrelated and make accurate prediction.
\end{itemize}

\section{Experiments}
\begin{table*}[h]
	\centering
	\resizebox{\textwidth}{!}{
		\begin{tabular}{c||c|c|c||c|c|c}
			\hline \hline
			\multirow{2}{*}{Model} & \multicolumn{3}{|c||}{PeMSD7(M) (15/ 30/ 60 min)} & \multicolumn{3}{c}{PeMSD7(L) (15/ 30/ 60min)}\\ \cline{2-7}
			& MAE & MAPE (\%) & RMSE & MAE & MAPE (\%) & RMSE \\ \hline \hline
			HA & 4.01 & 10.61 & 7.20 & 4.60 & 12.50 & 8.05 \\ \hline
			LSVR & 2.49/ 3.46/ 4.94 & 5.91/ 8.42/ 12.41 & 4.55/ 6.44/ 9.08 & 2.69/ 3.85/ 4.79 & 6.27/ 9.48/ 12.42 & 4.88/ 7.10/ 8.72 \\ \hline
			FNN & 2.53/ 3.73/ 5.28 & 6.05/ 9.48/ 13.73 & 4.46/ 6.46/ 8.75 & 2.61/ 3.71/ 5.36 & 6.11/ 9.20/ 14.68 & 4.74/ 6.76/ 9.09 \\ \hline
			FC-LSTM & 3.57/ 3.92/ 4.16 & 8.60/ 9.55/ 10.10 & 6.20/ 7.03/ 7.51 & 4.36/ 4.51/ 4.66 & 11.10/ 11.41/ 11.69 & 7.68/ 7.94/ 8.20\\ \hline
			STGCN & 2.24/ 3.02/ 4.01 & 5.20/ 7.27/ 9.77 & 4.07/ 5.70/ 7.55 & 2.37/ 3.27/ 4.35 & 5.56/ 7.98/ 11.17 & 4.32/ 6.21/ 8.27\\ \hline
			DCRNN & 2.25/ 2.98/ 3.83 & 5.30/ 7.39/ 9.85 & 4.04/ 5.58/ 7.19 & 2.36/ 3.24/ 4.34 & 5.51/ 8.18/ 11.91 & 4.45/ 6.31/ 8.33\\ \hline
			3D-TGCN &\textbf{2.23}/ \textbf{2.97}/ \textbf{3.65}& \textbf{5.13}/ \textbf{7.08}/ \textbf{8.79}& \textbf{3.93}/ \textbf{5.31}/ \textbf{6.66} & \textbf{2.27}/ \textbf{3.16}/ \textbf{3.79} & \textbf{5.31}/ \textbf{7.85}/ \textbf{9.76} & \textbf{4.18}/ \textbf{5.71}/ \textbf{7.13} \\ \hline

			 \hline
	\end{tabular}}
	\caption{Performance comparison of different models on PeMSD7 dataset.}
	\label{tab:pemsd7}
\end{table*}

\subsection{Datasets}
Our model is verified on two real-world traffic datasets which are used by two related state-of-the-art models: STGCN \cite{yuspatio} and DCRNN \cite{li2018diffusion}. 

\paragraph{PeMSD7} has a medium and a large scale \textbf{PeMSD7 (M)} and \textbf{PeMSD7 (L)} containing 228 and 1, 026 sensors separately among the District 7 of California. The data ranges from May and June of 2012 which are all at weekdays. 

\paragraph{PEMS-BAY} has 325 sensors in Bay Area and its collecting time is 6 months, ranging from Jan 2017 to June 2017.

These datasets are collected from California Transportation Agencies (Caltrans) Performance Measurement System (PeMS) in real-time by over 39, 000 sensor stations, which are deployed in the major metropolitan areas of California highway system\cite{chen2001freeway}. It is aggregated into 5-minute interval (228 time steps per day).
To compared strictly with those state-of-the-art models, we  follow all data preprocessing methods in each paper such as (1) the proportion and content of training, validation and test set, (2) utilizing the Gaussian kernel\cite{shuman2012emerging} to construct the spatial adjacency matrix.

\subsection{Experimental Settings and Baselines}
All experiments are compiled and tested on a Linux cluster(CPU: Intel(R) Xeon(R) CPU E5-2620 v4 @ 2.20GHz, GPU: Tesla P40). All model parameters are fine-tuned by gird search based on performance on validation set. Each prediction task uses past 60 minutes (i.e., 12 time steps are in time window $M=12$) to forecast traffic conditions in the next 15, 30 and 60 minutes ($H=3,6,12$).

\paragraph{Evaluation Metric}
Several criteria are introduced to evaluate 3D-TGCN, including the Mean Absolute Percentage Errors (MAPE), the Mean Absolute Errors (MAE) and the Root Mean Squared Errors (RMSE). All of them are used widely in traffic prediction tasks.

\textbf{3D-TGCN model}
The channels of each 3D graph convolutional layer in 3D-Conv block is 64. Receptive field of temporal graph $K$ is set to 3 and $K_t$ is set to 2. We use GLU as activation function in 3d-Conv block and sigmoid in output block. The learning rate  is set to $1e-2$ with a decay rate of $0.7$ after $3$ epochs. We train our models by minimizing the mean square error and mean absolute error using Adam for $30$ epochs with batch size as $50$. 

\paragraph{Baselines}
We compare our model with several baselines as follows:
\begin{itemize}
\item \textbf{HA}  Historical Average (HA), which treats the traffic speed value as a seasonal process and use weighted average of past several seasons as prediction value.
\item \textbf{SVR} Support Vector Regression (SVR), which uses linear support vector machine for regression tasks.
\item \textbf{FNN} Feed-Forward Neural Network (FNN), which is a classical neural network architecture with two hidden layers and loss function is RMSE.
\item \textbf{FC-LSTM} Full-Connected LSTM~\cite{sutskever2014sequence}, which is a Recurrent Neural Network with fully connected LSTM hidden units.
\item \textbf{DCRNN} Diffusion Convolutional Rrcurrent Neural Network(DCRNN)~\cite{li2018diffusion}, which models spatiotemporal dependencies with graph convolution into gated recurrent unit.
\item \textbf{STGCN} Spatio-Temporal Graph Convolutional Networks(STGCN)~\cite{yuspatio}, which models spatiotemporal dependencies with graph convolution into convolution structures.
\end{itemize}
All neural network based models are implemented in  Tensorflow~\cite{abadi2016tensorflow}.

\subsection{Experiment Results}
In this section, we compare our model with those baselines on the two datasets, shown in Table~\ref{tab:pemsd7} and \ref{tab:pems-bay}. It is obvious to observe that,  although all methods could perform well in short-term prediction, their performance varies greatly in long-term prediction. Deep learning models generally can achieve better performance than traditional machine learning models. Especially, STGCN and DCRNN, both of them have achieved significant improvement over other deep learning approaches since they extract additional information from spatial topology graph.  3D-TGCN could achieve the state-of-art performance especially when it only combines with temporal graph, demonstrating the importance of our proposed graph construction.

\paragraph{Accumulated Error of Sequence-to-Sequence Prediction}
RNN-based model and CNN model are different especially on the format of their output: while RNN-based ones conduct the next few time steps recursively, GCNs could predict few time steps recursively or directly predict the target time step.  Generally, RNN-based model performs better in time series tasks since the strategies such as  scheduled sampling~\cite{bengio2015scheduled} which can reduce accumulated error could be adopted on the sequence-to-sequence architecture. To compare these two types of outputs, we check the performance of 3D-TGCN: (1) predicting directly next $H$-th time step, (2) predicting the value of next $1, 2, \cdots, H$ time steps recursively. As we can see from Table \ref{tab:accumulated_error}, 3D-TGCN is more suitable for single step prediction task, it performs worse when predicting recursively due to accumulated error since its performance is close to DCRNN. However, 3D-TGCN could achieve better training efficiency since convolution-type models have less parameters than RNN-based models and STGCN.

\begin{table}
	\centering
	\resizebox{0.48\textwidth}{!}{
		\begin{tabular}{c||c|c|c}
			\hline \hline
			\multirow{2}{*}{Model} & \multicolumn{3}{|c}{PeMSD7(M) (15/ 30/ 60 min)} \\ \cline{2-4}
			& MAE & MAPE (\%) & RMSE \\ \hline \hline
			DCRNN& 2.25/ 2.98/ 3.83 & 5.30/ 7.39/ 9.85 & 4.04/ 5.58/ 7.19 \\ \hline
			STGCN & 2.24/ 3.02/ 4.01 & 5.20/ 7.27/ 9.77 & 4.07/ 5.70/ 7.55\\ \hline
			3D-TGCN (iteration)& 2.25/ 2.97/ 3.77 & 5.17/ 7.10 9.05 & 4.06/ 5.59/ 7.19 \\ \hline
			3D-TGCN (straightly) & 2.23/ 2.97/ 3.65 & 5.13/ 7.08/ 8.79& 3.93/ 5.31/ 6.66 \\ \hline
	\end{tabular}}
	\caption{Performance comparison of iteration / no iteration.}
	\label{tab:accumulated_error}
\end{table}

\begin{table}
	\centering
	\resizebox{0.48\textwidth}{!}{
		\begin{tabular}{c||c|c|c}
			\hline \hline
			\multirow{2}{*}{Model} & \multicolumn{3}{|c}{PeMSD7(M) (15/ 30/ 60 min)} \\ \cline{2-4}
			& MAE & MAPE (\%) & RMSE \\ \hline \hline
			STGCN (spatial) & 2.24/ 3.02/ 4.01 & 5.20/ 7.27/ 9.77 & 4.07/ 5.70/ 7.55\\ \hline
			STGCN (temporal) & 2.24/ 3.02/ 3.92 & 5.19/ 7.13/ 9.29 & 4.06/ 5.61/ 7.15 \\ \hline
			DCRNN (spatial) & 2.25/ 2.98/ 3.83 & 5.30/ 7.39/ 9.85 & 4.04/ 5.58/ 7.19 \\ \hline
			DCRNN (temporal) & 2.26/ 2.98/ 3.66 & 5.33/ 7.33/ 9.27 & 4.04/ 5.50/ 6.73 \\ \hline
			TGCN (spatial) &  2.24/ 3.00/ 3.76 & 5.21/ 7.12/ 8.96 &  3.96/ 5.37/ \textbf{6.64}\\ \hline
			TGCN (temporal) & \textbf{2.23}/ \textbf{2.97}/ \textbf{3.65} & \textbf{5.13}/ \textbf{7.08}/ \textbf{8.79}& \textbf{3.93}/ \textbf{5.31}/ 6.66 \\ \hline
	\end{tabular}}
	\caption{Performance comparison of spatial and temporal matrices.}
	\label{tab:t_and_s}
\end{table}

\begin{table}
	\centering
	\resizebox{0.48\textwidth}{!}{
		\begin{tabular}{c||c|c|c}
			\hline \hline
			\multirow{2}{*}{Model} & \multicolumn{3}{|c}{PEMS-BAY (15/ 30/ 60 min)} \\ \cline{2-4}
			& MAE & MAPE (\%) & RMSE \\ \hline \hline
			HA & 2.88 & 6.80 & 5.59 \\ \hline
			SVR & 1.85/ 2.48/ 3.28 & 3.80/ 5.50/ 8.00 & 3.59/ 5.18/ 7.08\\ \hline
			FNN & 1.49/ 2.04/ 2.88 & 3.09/ 4.59/ 7.11 & 3.25/ 4.45/ 5.99 \\ \hline
			FC-LSTM & 2.20/ 2.34/ 2.55 & 4.85/ 5.30/ 5.84 & 4.28/ 4.74/ 5.31 \\ \hline
			STGCN & 1.41/ 1.84/ 2.37 & 3.02/ 4.19/ 5.39 & 3.02/ 4.19/ 5.27 \\ \hline
			DCRNN & 1.38/ 1.74/ 2.07 & 2.9/ 3.9/ 4.9 & 2.95/ 3.97/ 4.74\\ \hline
			3D-TGCN & \textbf{1.34}/ \textbf{1.69}/ \textbf{2.07} & \textbf{2.78}/ \textbf{3.76}/ \textbf{4.76} & \textbf{2.79}/ \textbf{3.71}/ \textbf{4.56}\\ \hline
			\hline
	\end{tabular}}
	\caption{Performance comparison of different models on PEMS-BAY dataset.}
	\label{tab:pems-bay}
\end{table}

\paragraph{Temporal v.s. Spatial Pattern} 
Previous works  focus on incorporating spatial topology information of roads into time series prediction. Differently, our model switches to their dependencies of temporal patterns and has achieved the best performance on both short and long-term forecasting. The results of two types of graph construction are shown in Table~\ref{tab:t_and_s}. The performance of 3D-TGCN on dataset PeMSD7 is extremely well because road network of PeMSD7 is more complicated and systematic.

3D-TGCN does not require priori knowledge of spatial topology. On the contrary, it builds graphs on temporal dependency. As illustrated in Figure~\ref{fig:stg}, the left panel is spatial graph and right is temporal graph, the sparsity of them are both 5\%. The reason why temporal graph tends to be better than the spatial one is intuitive: (1) realistic data is full of noise, similar temporal dependency of different roads (maybe at distance) is much more important than spatial causality of neighbors; (2) traffic prediction is a time-series prediction task thus learning temporal pattern is more directly meaningful. 

An involuntary doubt about dynamic time warping is its computational complexity. Although $\mathcal{O}(n^2)$ is somewhat costly, in traffic prediction problem it is acceptable since the length of time series is 288 when time step is 5 min. Dataset PeMSD7(L) is one of biggest dataset in academic traffic speed field which has 1026 roads, in which the scalable version of DTW algorithm is still acceptable.

\section{Conclusion and Future works}
In this paper, we propose an original and effective deep learning framework 3D-TGCN for traffic prediction. It learns the relations between roads by comparing temporal similarity from the roads' times series and merges spatial and temporal information into 3D convolution simultaneously in the 3D graph convolutional layers. Numerical experiments show our model outperforms existing state-of-the-art models on two real-world datasets. Especially, our model does not require spatial topology. 3D-TGCN also achieves faster training  and better convergence. Our discovery of the new way of graph generation paves a  promising way for future graph-based learning approaches, due to the no need for spatial information based adjacency matrix, which in many cases are difficult to generate or achieve.  

\appendix

\bibliographystyle{named}
\bibliography{gcn3d}

\begin{thebibliography}{}

\bibitem[\protect\citeauthoryear{Abadi \bgroup \em et al.\egroup
  }{2016}]{abadi2016tensorflow}
Mart{\'\i}n Abadi, Paul Barham, Jianmin Chen, Zhifeng Chen, Andy Davis, Jeffrey
  Dean, Matthieu Devin, Sanjay Ghemawat, Geoffrey Irving, Michael Isard, et~al.
\newblock Tensorflow: A system for large-scale machine learning.
\newblock In {\em 12th $\{$USENIX$\}$ Symposium on Operating Systems Design and
  Implementation ($\{$OSDI$\}$ 16)}, pages 265--283, 2016.

\bibitem[\protect\citeauthoryear{Bengio \bgroup \em et al.\egroup
  }{2015}]{bengio2015scheduled}
Samy Bengio, Oriol Vinyals, Navdeep Jaitly, and Noam Shazeer.
\newblock Scheduled sampling for sequence prediction with recurrent neural
  networks.
\newblock In {\em Advances in Neural Information Processing Systems}, pages
  1171--1179, 2015.

\bibitem[\protect\citeauthoryear{Berndt and Clifford}{1994}]{berndt1994using}
Donald~J Berndt and James Clifford.
\newblock Using dynamic time warping to find patterns in time series.
\newblock In {\em KDD workshop}, volume~10, pages 359--370. Seattle, WA, 1994.

\bibitem[\protect\citeauthoryear{Bruna \bgroup \em et al.\egroup
  }{2013}]{bruna2013spectral}
Joan Bruna, Wojciech Zaremba, Arthur Szlam, and Yann LeCun.
\newblock Spectral networks and locally connected networks on graphs.
\newblock {\em arXiv preprint arXiv:1312.6203}, 2013.

\bibitem[\protect\citeauthoryear{Castro \bgroup \em et al.\egroup
  }{2012}]{castro2012urban}
Pablo~Samuel Castro, Daqing Zhang, and Shijian Li.
\newblock Urban traffic modelling and prediction using large scale taxi gps
  traces.
\newblock In {\em International Conference on Pervasive Computing}, pages
  57--72. Springer, 2012.

\bibitem[\protect\citeauthoryear{Chen \bgroup \em et al.\egroup
  }{2001}]{chen2001freeway}
Chao Chen, Karl Petty, Alexander Skabardonis, Pravin Varaiya, and Zhanfeng Jia.
\newblock Freeway performance measurement system: mining loop detector data.
\newblock {\em Transportation Research Record}, 1748(1):96--102, 2001.

\bibitem[\protect\citeauthoryear{Defferrard \bgroup \em et al.\egroup
  }{2016}]{defferrard2016convolutional}
Micha{\"e}l Defferrard, Xavier Bresson, and Pierre Vandergheynst.
\newblock Convolutional neural networks on graphs with fast localized spectral
  filtering.
\newblock In {\em Advances in Neural Information Processing Systems}, pages
  3844--3852, 2016.

\bibitem[\protect\citeauthoryear{Fitzpatrick \bgroup \em et al.\egroup
  }{2000}]{fitzpatrick2000speed}
Kay Fitzpatrick, Lily Elefteriadou, Douglas~W Harwood, JM~Collins, J~McFadden,
  Ingrid~B Anderson, Raymond~A Krammes, Nelson Irizarry, Kelly~D Parma, Karin~M
  Bauer, et~al.
\newblock Speed prediction for two-lane rural highways.
\newblock Technical report, 2000.

\bibitem[\protect\citeauthoryear{Haklay and
  Weber}{2008}]{haklay2008openstreetmap}
Mordechai Haklay and Patrick Weber.
\newblock Openstreetmap: User-generated street maps.
\newblock {\em IEEE Pervasive Computing}, 7(4):12--18, 2008.

\bibitem[\protect\citeauthoryear{Hammond \bgroup \em et al.\egroup
  }{2011}]{hammond2011wavelets}
David~K Hammond, Pierre Vandergheynst, and R{\'e}mi Gribonval.
\newblock Wavelets on graphs via spectral graph theory.
\newblock {\em Applied and Computational Harmonic Analysis}, 30(2):129--150,
  2011.

\bibitem[\protect\citeauthoryear{Hong}{2011}]{hong2011traffic}
Wei-Chiang Hong.
\newblock Traffic flow forecasting by seasonal svr with chaotic simulated
  annealing algorithm.
\newblock {\em Neurocomputing}, 74(12-13):2096--2107, 2011.

\bibitem[\protect\citeauthoryear{Kriegel \bgroup \em et al.\egroup
  }{2008}]{kriegel2008statistical}
Hans-Peter Kriegel, Matthias Renz, Matthias Schubert, and Andreas Zuefle.
\newblock Statistical density prediction in traffic networks.
\newblock In {\em Proceedings of the 2008 SIAM International Conference on Data
  Mining}, pages 692--703. SIAM, 2008.

\bibitem[\protect\citeauthoryear{Li \bgroup \em et al.\egroup
  }{2018}]{li2018diffusion}
Yaguang Li, Rose Yu, Cyrus Shahabi, and Yan Liu.
\newblock Diffusion convolutional recurrent neural network: Data-driven traffic
  forecasting.
\newblock 2018.

\bibitem[\protect\citeauthoryear{Ma \bgroup \em et al.\egroup
  }{2015}]{ma2015long}
Xiaolei Ma, Zhimin Tao, Yinhai Wang, Haiyang Yu, and Yunpeng Wang.
\newblock Long short-term memory neural network for traffic speed prediction
  using remote microwave sensor data.
\newblock {\em Transportation Research Part C: Emerging Technologies},
  54:187--197, 2015.

\bibitem[\protect\citeauthoryear{Ma \bgroup \em et al.\egroup
  }{2017}]{ma2017learning}
Xiaolei Ma, Zhuang Dai, Zhengbing He, Jihui Ma, Yong Wang, and Yunpeng Wang.
\newblock Learning traffic as images: a deep convolutional neural network for
  large-scale transportation network speed prediction.
\newblock {\em Sensors}, 17(4):818, 2017.

\bibitem[\protect\citeauthoryear{Niepert \bgroup \em et al.\egroup
  }{2016}]{niepert2016learning}
Mathias Niepert, Mohamed Ahmed, and Konstantin Kutzkov.
\newblock Learning convolutional neural networks for graphs.
\newblock In {\em International conference on machine learning}, pages
  2014--2023, 2016.

\bibitem[\protect\citeauthoryear{Okutani and
  Stephanedes}{1984}]{okutani1984dynamic}
Iwao Okutani and Yorgos~J Stephanedes.
\newblock Dynamic prediction of traffic volume through kalman filtering theory.
\newblock {\em Transportation Research Part B: Methodological}, 18(1):1--11,
  1984.

\bibitem[\protect\citeauthoryear{Ostring and
  Sirisena}{2001}]{ostring2001influence}
Sven~AM Ostring and Harsha Sirisena.
\newblock The influence of long-range dependence on traffic prediction.
\newblock In {\em ICC 2001. IEEE International Conference on Communications.
  Conference Record (Cat. No. 01CH37240)}, volume~4, pages 1000--1005. IEEE,
  2001.

\bibitem[\protect\citeauthoryear{Povinelli \bgroup \em et al.\egroup
  }{2004}]{povinelli2004time}
Richard~J Povinelli, Michael~T Johnson, Andrew~C Lindgren, and Jinjin Ye.
\newblock Time series classification using gaussian mixture models of
  reconstructed phase spaces.
\newblock {\em IEEE Transactions on Knowledge and Data Engineering},
  16(6):779--783, 2004.

\bibitem[\protect\citeauthoryear{Shuman \bgroup \em et al.\egroup
  }{2012}]{shuman2012emerging}
David~I Shuman, Sunil~K Narang, Pascal Frossard, Antonio Ortega, and Pierre
  Vandergheynst.
\newblock The emerging field of signal processing on graphs: Extending
  high-dimensional data analysis to networks and other irregular domains.
\newblock {\em arXiv preprint arXiv:1211.0053}, 2012.

\bibitem[\protect\citeauthoryear{Stathopoulos and
  Karlaftis}{2003}]{stathopoulos2003multivariate}
Anthony Stathopoulos and Matthew~G Karlaftis.
\newblock A multivariate state space approach for urban traffic flow modeling
  and prediction.
\newblock {\em Transportation Research Part C: Emerging Technologies},
  11(2):121--135, 2003.

\bibitem[\protect\citeauthoryear{Sutskever \bgroup \em et al.\egroup
  }{2014}]{sutskever2014sequence}
Ilya Sutskever, Oriol Vinyals, and Quoc~V Le.
\newblock Sequence to sequence learning with neural networks.
\newblock In {\em Advances in neural information processing systems}, pages
  3104--3112, 2014.

\bibitem[\protect\citeauthoryear{Vlahogianni \bgroup \em et al.\egroup
  }{2005}]{vlahogianni2005optimized}
Eleni~I Vlahogianni, Matthew~G Karlaftis, and John~C Golias.
\newblock Optimized and meta-optimized neural networks for short-term traffic
  flow prediction: A genetic approach.
\newblock {\em Transportation Research Part C: Emerging Technologies},
  13(3):211--234, 2005.

\bibitem[\protect\citeauthoryear{Vlahogianni}{2015}]{vlahogianni2015computtionl}
Eleni~I Vlahogianni.
\newblock Computtionl intelligence nd optimiztion for trnsporttion big dt:
  Chllenges nd opportunities.
\newblock In {\em Engineering and Applied Sciences Optimization}, pages
  107--128. Springer, 2015.

\bibitem[\protect\citeauthoryear{Williams and
  Hoel}{2003}]{williams2003modeling}
Billy~M Williams and Lester~A Hoel.
\newblock Modeling and forecasting vehicular traffic flow as a seasonal arima
  process: Theoretical basis and empirical results.
\newblock {\em Journal of transportation engineering}, 129(6):664--672, 2003.

\bibitem[\protect\citeauthoryear{Wu and Tan}{2016}]{wu2016short}
Yuankai Wu and Huachun Tan.
\newblock Short-term traffic flow forecasting with spatial-temporal correlation
  in a hybrid deep learning framework.
\newblock {\em arXiv preprint arXiv:1612.01022}, 2016.

\bibitem[\protect\citeauthoryear{Yu \bgroup \em et al.\egroup }{}]{yuspatio}
Bing Yu, Haoteng Yin, and Zhanxing Zhu.
\newblock Spatio-temporal graph convolutional networks: A deep learning
  framework for traffic forecasting.

\bibitem[\protect\citeauthoryear{Yu \bgroup \em et al.\egroup
  }{2004}]{yu2004switching}
Guoqiang Yu, Changshui Zhang, et~al.
\newblock Switching arima model based forecasting for traffic flow.
\newblock In {\em 2004 IEEE International Conference on Acoustics, Speech, and
  Signal Processing}, pages ii--429. IEEE, 2004.

\bibitem[\protect\citeauthoryear{Zhao \bgroup \em et al.\egroup
  }{2017}]{zhao2017lstm}
Zheng Zhao, Weihai Chen, Xingming Wu, Peter~CY Chen, and Jingmeng Liu.
\newblock Lstm network: a deep learning approach for short-term traffic
  forecast.
\newblock {\em IET Intelligent Transport Systems}, 11(2):68--75, 2017.

\end{thebibliography}

\end{document}